\documentclass[10pt,twocolumn,letterpaper]{article}

\usepackage[pagenumbers]{cvpr} 

\usepackage{amssymb}
\usepackage{pifont}
\newcommand{\xmark}{\ding{55}}%

\usepackage[dvipsnames]{xcolor}

\newcommand{\tb}[1]{\textbf{#1}}
\newcommand{\refmain}[1]{\textcolor{magenta}{#1}}
\newcommand{\envelope}{\ding{41}}

\definecolor{cvprblue}{rgb}{0.21,0.49,0.74}
\usepackage[pagebackref,breaklinks,colorlinks,citecolor=cvprblue]{hyperref}

\def\paperID{3489} 
\def\confName{CVPR}
\def\confYear{2024}
\newcommand{\name}{YOLO-World}

\title{\name{}: Real-Time Open-Vocabulary Object Detection}

\author{
Tianheng Cheng$^{3,2,*}$,~
Lin Song$^{1,*,\text{\envelope}}$,~
Yixiao Ge$^{1,2,\dagger}$,~
Wenyu Liu$^{3}$,~
Xinggang Wang$^{3,\text{\envelope}}$,~
Ying Shan$^{1,2}$ \\
\textbf{\small $^*$equal contribution}\quad \textbf{\small $^\dagger$ project lead}\quad \textbf{\small $^\text{\envelope}$ corresponding author}\\~\\
$^1$~Tencent AI Lab~
$^2$~ARC Lab, Tencent PCG\\
$^3$~School of EIC, Huazhong University of Science \& Technology
\\~\\
Code \& Models: \href{https://github.com/AILab-CVC/YOLO-World}{YOLO-World}
}

\begin{document}
\maketitle

\begin{abstract}
The You Only Look Once (YOLO) series of detectors have established themselves as efficient and practical tools. However, their reliance on predefined and trained object categories limits their applicability in open scenarios. Addressing this limitation, we introduce \name{}, an innovative approach that enhances YOLO with open-vocabulary detection capabilities through vision-language modeling and pre-training on large-scale datasets. Specifically, we propose a new Re-parameterizable Vision-Language Path Aggregation Network (RepVL-PAN) and region-text contrastive loss to facilitate the interaction between visual and linguistic information. Our method excels in detecting a wide range of objects in a zero-shot manner with high efficiency. On the challenging LVIS dataset, \name{} achieves 35.4 AP with 52.0 FPS on V100, which outperforms many state-of-the-art methods in terms of both accuracy and speed. Furthermore, the fine-tuned \name{} achieves remarkable performance on several downstream tasks, including object detection and open-vocabulary instance segmentation.

\end{abstract}


\section{Introduction}
\label{sec:intro}

Object detection has been a long-standing and fundamental challenge in computer vision with numerous applications in image understanding, robotics, and autonomous vehicles.
Tremendous works~\cite{ren2015faster,YOLOv1,FPN,MaskR-CNN} have achieved significant breakthroughs in object detection with the development of deep neural networks.
Despite the success of these methods, they remain limited as they only handle object detection with a fixed vocabulary, \eg, 80 categories in the COCO~\cite{COCO} dataset.
Once object categories are defined and labeled, trained detectors can only detect those specific categories, thus limiting the ability and applicability of open scenarios.

\begin{figure}
    \centering
    \includegraphics[width=\linewidth]{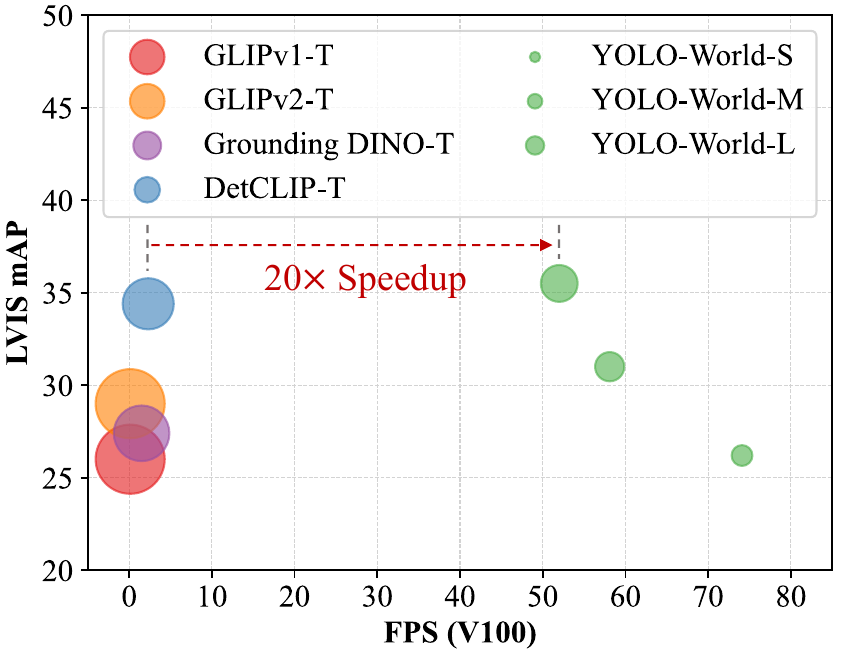}
    \caption{\textbf{Speed-and-Accuracy Curve.} We compare \name{} with recent open-vocabulary methods in terms of speed and accuracy. All models are evaluated on the LVIS \texttt{minival} and inference speeds are measured on one NVIDIA V100 w/o TensorRT. The size of the circle represents the model's size.}
    \vspace{-13pt}
    \label{fig:speed_acc}
\end{figure}

Recent works~\cite{OVD,ViLD,DetPro,BARON,EDADet} have explored the prevalent vision-language models~\cite{CLIP,ALIGN} to address open-vocabulary detection~\cite{OVD} through distilling vocabulary knowledge from language encoders, \eg, BERT~\cite{BERT}.
However, these distillation-based methods are much limited due to the scarcity of training data with a limited diversity of vocabulary, \eg, OV-COCO~\cite{OVD} containing 48 base categories.
Several methods~\cite{GLIP,GLIPv2,GroundingDINO,DetCLIP,DetCLIPv2} reformulate object detection training as region-level vision-language pre-training and train open-vocabulary object detectors at scale.
However, those methods still struggle for detection in real-world scenarios, which suffer from two aspects: (1) heavy computation burden and (2) complicated deployment for edge devices.
Previous works~\cite{GLIP,GLIPv2,GroundingDINO,DetCLIP,DetCLIPv2} have demonstrated the promising performance of pre-training large detectors while pre-training small detectors to endow them with open recognition capabilities remains unexplored.

\begin{figure*}
    \centering
    \includegraphics[width=\linewidth]{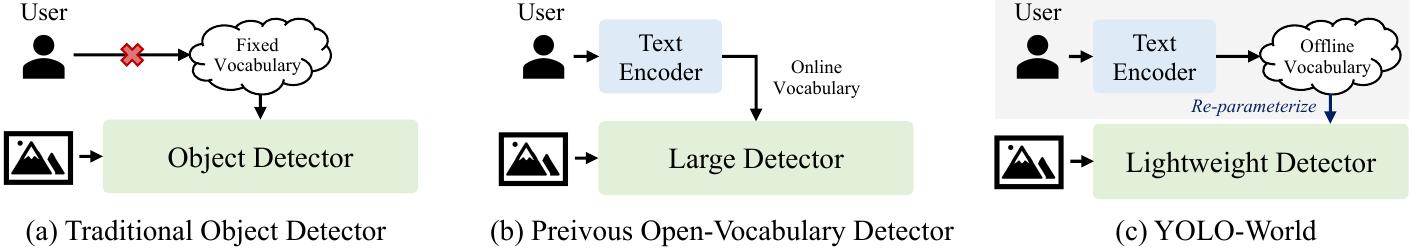}
    \caption{\textbf{Comparison with Detection Paradigms.} \textbf{(a) Traditional Object Detector}: These object detectors can only detect objects within the fixed vocabulary pre-defined by the training datasets, \eg, 80 categories of COCO dataset~\cite{COCO}. The fixed vocabulary limits the extension for open scenes. 
    \textbf{(b) Previous Open-Vocabulary Detectors:} Previous methods tend to develop large and heavy detectors for open-vocabulary detection which intuitively have strong capacity. In addition, these detectors simultaneously encode images and texts as input for prediction, which is time-consuming for practical applications.
    \textbf{(c) \name{}:} We demonstrate the strong open-vocabulary performance of lightweight detectors, \eg, YOLO detectors~\cite{YOLO,yolov8_ultralytics}, which is of great significance for real-world applications. Rather than using online vocabulary, we present a \textit{prompt-then-detect} paradigm for efficient inference, in which the user generates a series of prompts according to the need and the prompts will be encoded into an offline vocabulary. Then it can be re-parameterized as the model weights for deployment and further acceleration.}
    \label{fig:paramdigm_comparison}
    \vspace{-10pt}
\end{figure*}

In this paper, we present \name{}, aiming for high-efficiency open-vocabulary object detection, and explore large-scale pre-training schemes to boost the traditional YOLO detectors to a new open-vocabulary world.
Compared to previous methods, the proposed \name{} is remarkably efficient with high inference speed and easy to deploy for downstream applications.
Specifically, \name{} follows the standard YOLO architecture~\cite{yolov8_ultralytics} and leverages the pre-trained CLIP~\cite{CLIP} text encoder to encode the input texts.
We further propose the Re-parameterizable Vision-Language Path Aggregation Network (RepVL-PAN) to connect text features and image features for better visual-semantic representation.
During inference, the text encoder can be removed and the text embeddings can be re-parameterized into weights of RepVL-PAN for efficient deployment.
We further investigate the open-vocabulary pre-training scheme for YOLO detectors through region-text contrastive learning on large-scale datasets, which unifies detection data, grounding data, and image-text data into region-text pairs.
The pre-trained \name{} with abundant region-text pairs demonstrates a strong capability for large vocabulary detection and training more data leads to greater improvements in open-vocabulary capability.

In addition, we explore a \textit{prompt-then-detect} paradigm to further improve the efficiency of open-vocabulary object detection in real-world scenarios.
As illustrated in Fig.~\ref{fig:paramdigm_comparison}, traditional object detectors~\cite{MaskR-CNN,YOLO,YOLOv1,Yolov3,YOLOv6,YOLOv7,yolov8_ultralytics} concentrate on the fixed-vocabulary (close-set) detection with predefined and trained categories.
While previous open-vocabulary detectors~\cite{GLIP,GLIPv2,GroundingDINO,DetCLIP} encode the prompts of a user for online vocabulary with text encoders and detect objects.
Notably, those methods tend to employ large detectors with heavy backbones, \eg, Swin-L~\cite{SwinTransformer}, to increase the open-vocabulary capacity.
In contrast, the \textit{prompt-then-detect} paradigm (Fig.~\ref{fig:paramdigm_comparison} (c)) first encodes the prompts of a user to build an offline vocabulary and the vocabulary varies with different needs.
Then, the efficient detector can infer the offline vocabulary on the fly without re-encoding the prompts.
For practical applications, once we have trained the detector, \ie, \name{}, we can pre-encode the prompts or categories to build an offline vocabulary and then seamlessly integrate it into the detector.


Our main contributions can be summarized into three folds:
\begin{itemize}
    \item We introduce the \name{}, a cutting-edge open-vocabulary object detector with high efficiency for real-world applications.
    \item We propose a Re-parameterizable Vision-Language PAN to connect vision and language features and an open-vocabulary region-text contrastive pre-training scheme for \name{}.
    \item The proposed \name{} pre-trained on large-scale datasets demonstrates strong zero-shot performance and achieves 35.4 AP on LVIS with 52.0 FPS.
    The pre-trained \name{} can be easily adapted to downstream tasks, \eg, open-vocabulary instance segmentation and referring object detection.
    Moreover, the pre-trained weights and codes of \name{} will be open-sourced to facilitate more practical applications. 
\end{itemize}

\begin{figure*}[]
    \centering
    \includegraphics[width=\linewidth]{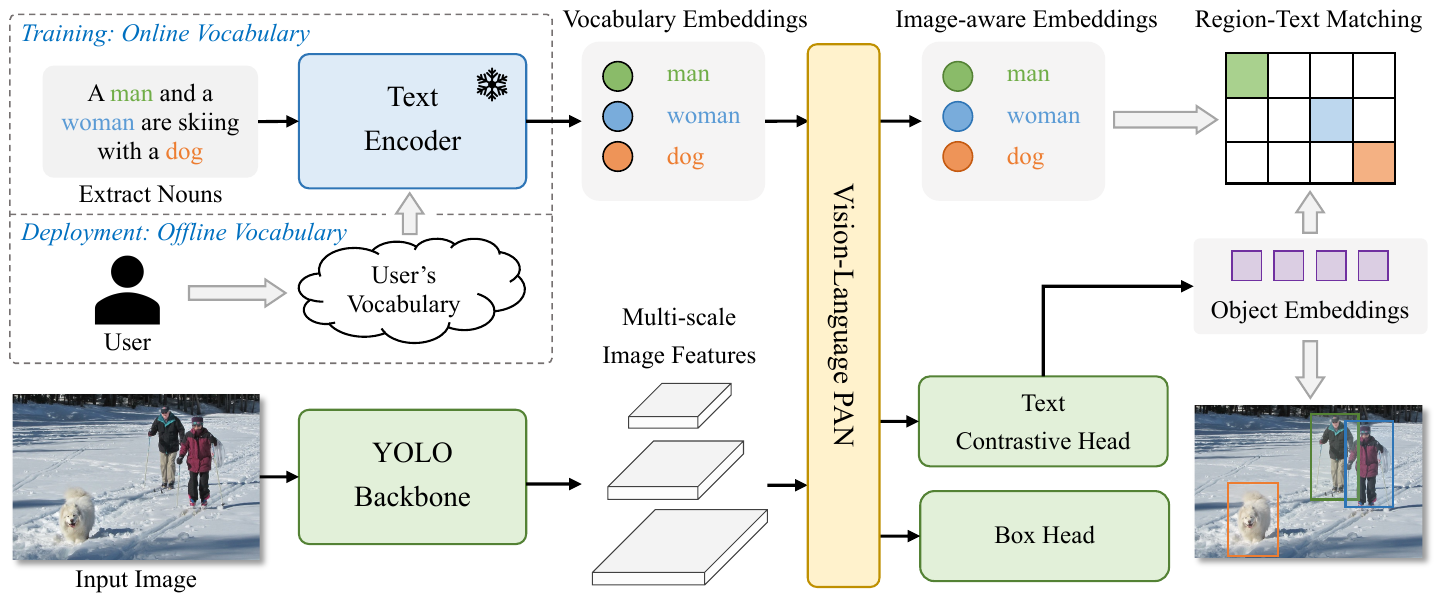}
    \caption{\textbf{Overall Architecture of \name{}.} Compared to traditional YOLO detectors, \name{} as an open-vocabulary detector adopts text as input. The \textit{Text Encoder}  first encodes the input text input text embeddings. Then the \textit{Image Encoder} encodes the input image into multi-scale image features and the proposed \textit{RepVL-PAN} exploits the multi-level cross-modality fusion for both image and text features. Finally, \name{} predicts the regressed bounding boxes and the object embeddings for matching the categories or nouns that appeared in the input text.}
    \vspace{-10pt}
    \label{fig:main_figure}
\end{figure*}

\label{sec:method}
\section{Related Works}
\label{sec:related_works}
\subsection{Traditional Object Detection}
Prevalent object detection research concentrates on fixed-vocabulary (close-set) detection, in which object detectors are trained on datasets with pre-defined categories, \eg, COCO dataset~\cite{COCO} and Objects365 dataset~\cite{Shao2019Objects365AL}, and then detect objects within the fixed set of categories.
During the past decades, the methods for traditional object detection can be simply categorized into three groups, \ie, region-based methods, pixel-based methods, and query-based methods.
The region-based methods~\cite{RCNN,FastRCNN,FasterRCNN,FPN,MaskR-CNN}, such as Faster R-CNN~\cite{FasterRCNN}, adopt a two-stage framework for proposal generation~\cite{FasterRCNN} and RoI-wise (Region-of-Interest) classification and regression.
The pixel-based methods~\cite{YOLO,SSD,RetinaNet,FCOS,ATSS} tend to be one-stage detectors, which perform classification and regression over pr\textbf{}e-defined anchors or pixels.
DETR~\cite{DETR} first explores object detection through transformers~\cite{Transformer} and inspires extensive query-based methods~\cite{DeformableDETR}.
In terms of inference speed, Redmon~\etal presents YOLOs~\cite{YOLO,YOLO9000,Yolov3} which exploit simple convolutional architectures for real-time object detection.
Several works~\cite{YOLOv6,YOLOv7,YOLOX,PPYOLO,PPYOLOE} propose various architectures or designs for YOLO, including path aggregation networks~\cite{PAFPN}, cross-stage partial networks~\cite{CSPNet}, and re-parameterization~\cite{RepVGG}, which further improve both speed and accuracy.
In comparison to previous YOLOs, \name{} in this paper aims to detect objects beyond the fixed vocabulary with strong generalization ability.

\subsection{Open-Vocabulary Object Detection}
Open-vocabulary object detection (OVD)~\cite{OVD} has emerged as a new trend for modern object detection, which aims to detect objects beyond the predefined categories. Early works~\cite{ViLD} follow the standard OVD setting~\cite{OVD} by training detectors on the base classes and evaluating the novel (unknown) classes. Nevertheless, this open-vocabulary setting can evaluate the capability of detectors to detect and recognize novel objects, it is still limited for open scenarios and lacks generalization ability to other domains due to training on the limited dataset and vocabulary. Inspired by vision-language pre-training~\cite{CLIP,ALIGN}, recent works~\cite{RegionCLIP,Detic,FVLM,DetPro,BARON} formulate open-vocabulary object detection as image-text matching and exploit large-scale image-text data to increase the training vocabulary at scale.
OWL-ViTs~\cite{owlv1,owlv2} fine-tune the simple vision transformers~\cite{ViT} with detection and grounding datasets and build the simple open-vocabulary detectors with promising performance.
GLIP~\cite{GLIP} presents a pre-training framework for open-vocabulary detection based on phrase grounding and evaluates in a zero-shot setting. Grounding DINO~\cite{GroundingDINO} incorporates the grounded pre-training~\cite{GLIP} into detection transformers~\cite{DINO} with cross-modality fusions.
Several methods~\cite{GLIPv2,VLDet,DetCLIP,DetCLIPv2} unify detection datasets and image-text datasets through region-text matching and pre-train detectors with large-scale image-text pairs, achieving promising performance and generalization. However, these methods often use heavy detectors like ATSS~\cite{ATSS} or DINO~\cite{DINO} with Swin-L~\cite{SwinTransformer} as a backbone, leading to high computational demands and deployment challenges. In contrast, we present \name{}, aiming for efficient open-vocabulary object detection with real-time inference and easier downstream application deployment. Differing from ZSD-YOLO~\cite{ZSD_YOLO}, which also explores open-vocabulary detection~\cite{OVD} with YOLO through language model alignment, \name{} introduces a novel YOLO framework with an effective pre-training strategy, enhancing open-vocabulary performance and generalization.

\section{Method}

\subsection{Pre-training Formulation: Region-Text Pairs}
The traditional object detection methods, including the YOLO-series~\cite{yolov8_ultralytics}, are trained with instance annotations $\Omega=\{B_i, c_i\}_{i=1}^N$, which consist of bounding boxes $\{B_i\}$ and category labels $\{c_i\}$.
In this paper, we reformulate the instance annotations as region-text pairs $\Omega=\{B_i, t_i\}_{i=1}^N$, where $t_i$ is the corresponding text for the region $B_i$.
Specifically, the text $t_i$ can be the category name, noun phrases, or object descriptions.
Moreover, \name{} adopts both the image $I$ and texts $T$ (a set of nouns) as input and outputs predicted boxes $\{\hat{B}_k\}$ and the corresponding object embeddings $\{e_k\}$ ($e_k \in \mathbb{R}^D$).

\subsection{Model Architecture}
The overall architecture of the proposed \name{} is illustrated in Fig.~\ref{fig:main_figure}, which consists of a \textit{YOLO detector}, a \textit{Text Encoder}, and a \textit{Re-parameterizable Vision-Language Path Aggregation Network} (RepVL-PAN).
Given the input text, the text encoder in \name{} encodes the text into text embeddings.
The image encoder in the YOLO detector extracts the multi-scale features from the input image.
Then we leverage the RepVL-PAN to enhance both text and image representation by exploiting the cross-modality fusion between image features and text embeddings.

\paragraph{YOLO Detector.}
\name{} is mainly developed based on YOLOv8~\cite{yolov8_ultralytics}, which contains a Darknet backbone~\cite{YOLOv1, yolov8_ultralytics} as the image encoder, a path aggregation network (PAN) for multi-scale feature pyramids, and a head for bounding box regression and object embeddings.

\paragraph{Text Encoder.}
Given the text $T$, we adopt the Transformer text encoder pre-trained by CLIP~\cite{CLIP} to extract the corresponding text embeddings $W\!=\!\texttt{TextEncoder}(T) \!\in\! \mathbb{R}^{C\!\times\!D}$, where $C$ is the number of nouns and $D$ is the embedding dimension.
The CLIP text encoder offers better visual-semantic capabilities for connecting visual objects with texts compared to text-only language encoders~\cite{BERT}.
When the input text is a caption or referring expression, we adopt the simple n-gram algorithm to extract the noun phrases and then feed them into the text encoder.

\paragraph{Text Contrastive Head.}
Following previous works~\cite{yolov8_ultralytics}, we adopt the decoupled head with two $3\times3$ convs to regress bounding boxes $\{b_k\}^K_{k=1}$ and object embeddings $\{e_k\}^K_{k=1}$, where $K$ denotes the number of objects.
We present a text contrastive head to obtain the object-text similarity $s_{k,j}$ by:
\begin{equation}
    s_{k,j} = \alpha\cdot \texttt{L2-Norm}(e_k) \cdot \texttt{L2-Norm}(w_j)^{\top} + \beta,
\end{equation}
where $\texttt{L2-Norm}(\cdot)$ is the L2 normalization and $w_j \in W$ is the $j$-th text embeddings. In addition, we add the affine transformation with the learnable scaling factor $\alpha$ and shifting factor $\beta$.
Both the L2 norms and the affine transformations are important for stabilizing the region-text training.

\paragraph{Training with Online Vocabulary.}
During training, we construct an online vocabulary $T$ for each mosaic sample containing 4 images.
Specifically, we sample all positive nouns involved in the mosaic images and randomly sample some negative nouns from the corresponding dataset.
The vocabulary for each mosaic sample contains at most $M$ nouns, and $M$ is set to 80 as default.

\paragraph{Inference with Offline Vocabulary.}
At the inference stage, we present a \textit{prompt-then-detect} strategy with an offline vocabulary for further efficiency.
As shown in Fig.~\ref{fig:main_figure}, the user can define a series of custom prompts, which might include captions or categories.
We then utilize the text encoder to encode these prompts and obtain offline vocabulary embeddings.
The offline vocabulary allows for avoiding computation for each input and provides the flexibility to adjust the vocabulary as needed.

\subsection{Re-parameterizable Vision-Language PAN}
\begin{figure}
    \centering
    \includegraphics[width=\linewidth]{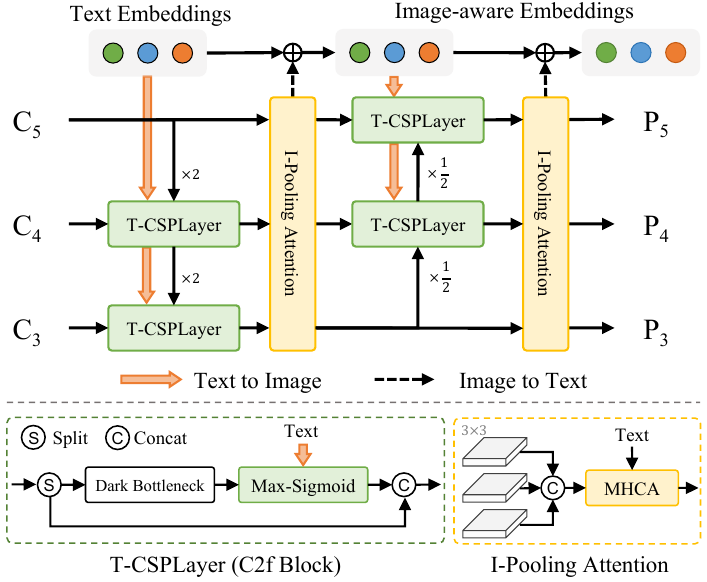}
    \caption{\textbf{Illustration of the RepVL-PAN.} The proposed RepVL-PAN adopts the \textit{Text-guided CSPLayer} (T-CSPLayer) for injecting language information into image features and the \textit{Image Pooling Attention} (I-Pooling Attention) for enhancing image-aware text embeddings.}
    \label{fig:vlpan}
    \vspace{-10pt}
\end{figure}

Fig.~\ref{fig:vlpan} shows the structure of the proposed RepVL-PAN which follows the top-down and bottom-up paths in \cite{PAFPN,yolov8_ultralytics} to establish the feature pyramids $\{P_3, P_4, P_5\}$ with the multi-scale image features $\{C_3, C_4, C_5\}$.
Furthermore, we propose the Text-guided CSPLayer (T-CSPLayer) and Image-Pooling Attention (I-Pooling Attention) to further enhance the interaction between image features and text features, which can improve the visual-semantic representation for open-vocabulary capability.
During inference, the offline vocabulary embeddings can be re-parameterized into weights of convolutional or linear layers for deployment.

\paragraph{Text-guided CSPLayer.}
As Fig.~\ref{fig:vlpan} illustrates, the cross-stage partial layers (CSPLayer) are utilized after the top-down or bottom-up fusion.
We extend the CSPLayer (also called \texttt{C2f}) of \cite{yolov8_ultralytics} by incorporating text guidance into multi-scale image features to form the Text-guided CSPLayer.
Specifically, given the text embeddings $W$ and image features $X_l \in \mathbb{R}^{H\times W\times D}$ ($l \in \{3,4,5\}$),
we adopt the \textit{max-sigmoid attention} after the last dark bottleneck block to aggregate text features into image features by:
\begin{equation}
X_l' = X_l\cdot\delta(\max_{j\in\{1..C\}}(X_lW_j^{\top}))^{\top},
\end{equation}
where the updated $X_l'$ is concatenated with the cross-stage features as output. The $\delta$ indicates the sigmoid function.

\paragraph{Image-Pooling Attention.}
To enhance the text embeddings with image-aware information, we aggregate image features to update the text embeddings by proposing the Image-Pooling Attention.
Rather than directly using cross-attention on image features, we leverage max pooling on multi-scale features to obtain $3\!\times\!3$ regions, resulting in a total of 27 patch tokens $\tilde{X}\in \mathbb{R}^{27\times D}$.
The text embeddings are then updated by:
\begin{equation}
    W' = W + \texttt{MultiHead-Attention}(W, \tilde{X}, \tilde{X})
\end{equation}

\subsection{Pre-training Schemes}
In this section, we present the training schemes for pre-training \name{} on large-scale detection, grounding, and image-text datasets.
\paragraph{Learning from Region-Text Contrastive Loss.}
Given the mosaic sample $I$ and texts $T$, \name{} outputs $K$ object predictions $\{B_k, s_k\}_{k=1}^K$ along with annotations $\Omega=\{B_i,t_i\}^N_{i=1}$.
We follow \cite{yolov8_ultralytics} and leverage task-aligned label assignment~\cite{TOOD} to match the predictions with ground-truth annotations and assign each positive prediction with a text index as the classification label. 
Based on this vocabulary, we construct the region-text contrastive loss $\mathcal{L}_{\text{con}}$ with region-text pairs through cross entropy between object-text (region-text) similarity and object-text assignments.
In addition, we adopt IoU loss and distributed focal loss for bounding box regression and the total training loss is defined as: $    \mathcal{L}(I) = \mathcal{L}_{\text{con}} + \lambda_I\cdot(\mathcal{L}_{\text{iou}} + \mathcal{L}_{\text{dfl}}),$
where $\lambda_I$ is an indicator factor and set to 1 when input image $I$ is from detection or grounding data and set to 0 when it is from the image-text data.
Considering image-text datasets have noisy boxes, we only calculate the regression loss for samples with accurate bounding boxes.

\paragraph{Pseudo Labeling with Image-Text Data.}
\label{sec:pseudo_labeling}
Rather than directly using image-text pairs for pre-training, we propose an automatic labeling approach to generate region-text pairs.
Specifically, the labeling approach contains three steps: 
(1) \textit{extract noun phrases}: we first utilize the n-gram algorithm to extract noun phrases from the text; 
(2) \textit{pseudo labeling}: we adopt a pre-trained open-vocabulary detector, \eg, GLIP~\cite{GLIP}, to generate pseudo boxes for the given noun phrases for each image, thus providing the coarse region-text pairs.
(3) \textit{filtering}: We employ the pre-trained CLIP~\cite{CLIP} to evaluate the relevance of image-text pairs and region-text pairs, and filter the low-relevance pseudo annotations and images. We further filter redundant bounding boxes by incorporating methods such as Non-Maximum Suppression (NMS).
We suggest the readers refer to the appendix for the detailed approach.
With the above approach, we sample and label 246k images from CC3M~\cite{CC3M} with 821k pseudo annotations.

    


\section{Experiments}
\label{sec:experiments}
In this section, we demonstrate the effectiveness of the proposed \name{} by pre-training it on large-scale datasets and evaluating \name{} in a zero-shot manner on both LVIS benchmark and COCO benchmark (Sec.~\ref{subsec:pretrain}).
We also evaluate the fine-tuning performance of \name{} on COCO, LVIS for object detection.

\subsection{Implementation Details}
The \name{} is developed based on the MMYOLO toolbox~\cite{mmyolo2022} and the MMDetection toolbox~\cite{mmdetection}.
Following \cite{yolov8_ultralytics}, we provide three variants of \name{} for different latency requirements, \eg, small (S), medium (M), and large (L).
We adopt the open-source CLIP~\cite{CLIP} text encoder with pre-trained weights to encode the input text.
Unless specified, we measure the inference speeds of all models on one NVIDIA V100 GPU without extra acceleration mechanisms, \eg, FP16 or TensorRT.

\subsection{Pre-training}
\label{subsec:pretrain}
\paragraph{Experimental Setup.}
At the pre-training stage, we adopt the AdamW optimizer~\cite{AdamW} with an initial learning rate of 0.002 and weight decay of 0.05.
\name{} is pre-trained for 100 epochs on on 32 NVIDIA V100 GPUs with a total batch size of 512.
During pre-training, we follow previous works~\cite{yolov8_ultralytics} and adopt color augmentation, random affine, random flip, and mosaic with 4 images for data augmentation.
The text encoder is frozen during pre-training.

\paragraph{Pre-training Data.}
For pre-training \name{}, we mainly adopt detection or grounding datasets including Objects365 (V1)~\cite{Shao2019Objects365AL}, GQA~\cite{GQA}, Flickr30k~\cite{Flickr}, as specified in Tab.~\ref{tab:pretrain_data}.
Following \cite{GLIP}, we exclude the images from the COCO dataset in GoldG~\cite{mdetr} (GQA and Flickr30k).
The annotations of the detection datasets used for pre-training contain both bounding boxes and categories or noun phrases.
In addition, we also extend the pre-training data with image-text pairs, \ie, CC3M$^{\dagger}$~\cite{CC3M}, which we have labeled 246k images through the pseudo-labeling method discussed in Sec.~\ref{sec:pseudo_labeling}.

\begin{table}
    \centering
    \renewcommand\arraystretch{1.2}
    \scalebox{0.88}{\begin{tabular}{llccc}
    \toprule
    Dataset & Type & Vocab. & Images & Anno. \\
    \hline
    Objects365V1~\cite{Shao2019Objects365AL} & Detection & 365 & 609k & 9,621k\\
    GQA~\cite{GQA} & Grounding & - & 621k & 3,681k \\
    Flickr~\cite{Flickr} & Grounding & - & 149k & 641k\\
    CC3M$\dagger$~\cite{CC3M} & Image-Text & - & 246k & 821k \\
    \bottomrule
    \end{tabular}}
    \caption{\textbf{Pre-training Data.} The specifications of the datasets used for pre-training \name{}.}
    \vspace{-10pt}
    \label{tab:pretrain_data}
\end{table}

\paragraph{Zero-shot Evaluation.}
After pre-training, we directly evaluate the proposed \name{} on the LVIS dataset~\cite{LVIS} in a zero-shot manner.
The LVIS dataset contains 1203 object categories, which is much more than the categories of the pre-training detection datasets and can measure the performance on large vocabulary detection.
Following previous works~\cite{mdetr,GLIP,DetCLIP,DetCLIPv2}, we mainly evaluate on LVIS \texttt{minival}~\cite{mdetr} and report the \textit{Fixed AP}~\cite{largevocabdevil} for comparison. The maximum number of predictions is set to 1000.

\paragraph{Main Results on LVIS Object Detection.}
In Tab.~\ref{tab:zero_shot_lvis}, we compare the proposed \name{} with recent state-of-the-art methods~\cite{mdetr,GLIPv2,DetCLIP,DetCLIPv2,GroundingDINO} on LVIS benchmark in a zero-shot manner.
Considering the computation burden and model parameters, we mainly compare with those methods based on lighter backbones, \eg, Swin-T~\cite{SwinTransformer}.
Remarkably, \name{} outperforms previous state-of-the-art methods in terms of zero-shot performance and inference speed.
Compared to GLIP, GLIPv2, and Grounding DINO, which incorporate more data, \eg, Cap4M (CC3M+SBU~\cite{SBU}), \name{} pre-trained on O365 \& GolG obtains better performance even with fewer model parameters.
Compared to DetCLIP, \name{} achieves comparable performance (35.4 \textit{v.s.} 34.4) while obtaining $20\times$ increase in inference speed.
\textit{The experimental results also demonstrate that small models, \eg, \name{}-S with 13M parameters, can be used for vision-language pre-training and obtain strong open-vocabulary capabilities.}

\begin{table*}[]
    \centering
    \renewcommand\arraystretch{1.2}
    \scalebox{0.95}{\begin{tabular}{llcl c cccc}
    \toprule
    Method & Backbone & Params & Pre-trained Data & FPS & AP & AP$_r$ & AP$_c$ & AP$_f$\\
    \hline
    MDETR~\cite{mdetr} & R-101~\cite{ResNet} & 169M & GoldG & - & 24.2 & 20.9 & 24.3 & 24.2 \\
    GLIP-T~\cite{GLIP} &Swin-T~\cite{SwinTransformer} & 232M & O365,GoldG & 0.12 &24.9 & 17.7 & 19.5 & 31.0 \\
    GLIP-T~\cite{GLIP} &Swin-T~\cite{SwinTransformer} & 232M & O365,GoldG,Cap4M & 0.12 & 26.0 & 20.8 & 21.4 & 31.0 \\
    GLIPv2-T~\cite{GLIPv2} & Swin-T~\cite{SwinTransformer}  & 232M & O365,GoldG & 0.12 & 26.9 & - & - & - \\
    GLIPv2-T~\cite{GLIPv2} & Swin-T~\cite{SwinTransformer} & 232M &  O365,GoldG,Cap4M & 0.12 & 29.0 & - & - & - \\
    Grounding DINO-T~\cite{GroundingDINO} & Swin-T~\cite{SwinTransformer} & 172M & O365,GoldG & 1.5 & 25.6 & 14.4 & 19.6 & 32.2 \\
    Grounding DINO-T~\cite{GroundingDINO} & Swin-T~\cite{SwinTransformer} & 172M & O365,GoldG,Cap4M & 1.5 & 27.4 & 18.1 & 23.3 & 32.7 \\
    DetCLIP-T~\cite{DetCLIP} & Swin-T~\cite{SwinTransformer} & 155M &  O365,GoldG & 2.3 & 34.4 & 26.9 & 33.9 & 36.3 \\
    \hline

    \name{}-S & YOLOv8-S & 13M (77M) & O365,GoldG & 74.1 (19.9) & 26.2 & 19.1 & 23.6 & 29.8 \\
    \name{}-M & YOLOv8-M & 29M (92M) & O365,GoldG & 58.1 (18.5) & 31.0 & 23.8 & 29.2 & 33.9 \\
    \name{}-L & YOLOv8-L & 48M (110M) & O365,GoldG & 52.0 (17.6) & 35.0 & 27.1 & 32.8 & 38.3 \\
    \name{}-L & YOLOv8-L & 48M (110M) & O365,GoldG,CC3M$^\dagger$ & 52.0 (17.6) & 35.4 & 27.6 & 34.1 & 38.0 \\

    \bottomrule
    \end{tabular}}
    \vspace{-4pt}
    \caption{\textbf{Zero-shot Evaluation on LVIS.} We evaluate \name{} on LVIS \texttt{minival}~\cite{mdetr} in a zero-shot manner. We report the \textit{Fixed AP}~\cite{largevocabdevil} for a fair comparison with recent methods. $^{\dagger}$ denotes the pseudo-labeled CC3M in our setting, which contains 246k samples. The FPS is evaluated on one NVIDIA V100 GPU w/o TensorRT. The parameters and FPS of \name{} are evaluated for both the re-parameterized version (w/o bracket) and the original version (w/ bracket).}
    \label{tab:zero_shot_lvis}
\end{table*}

\subsection{Ablation Experiments}
We provide extensive ablation studies to analyze \name{} from two primary aspects, \ie, pre-training and architecture.
Unless specified, we mainly conduct ablation experiments based on \name{}-L and pre-train Objects365 with zero-shot evaluation on LVIS \texttt{minival}.

\paragraph{Pre-training Data.}
In Tab.~\ref{tab:ablation_pretrain_data}, we evaluate the performance of pre-training \name{} using different data.
Compared to the baseline trained on Objects365, adding GQA can significantly improve performance with an 8.4 AP gain on LVIS.
This improvement can be attributed to the richer textual information provided by the GQA dataset, which can enhance the model's ability to recognize large vocabulary objects.
Adding part of CC3M samples (8\% of the full datasets) can further bring 0.5 AP gain with 1.3 AP on rare objects.
Tab.~\ref{tab:ablation_pretrain_data} demonstrates that adding more data can effectively improve the detection capabilities on large-vocabulary scenarios.
Furthermore, as the amount of data increases, the performance continues to improve, highlighting the benefits of leveraging larger and more diverse datasets for training.

\begin{table}[]
    \centering
    \renewcommand\arraystretch{1.2}
    \begin{tabular}{ll cccc}
    \toprule
    Pre-trained Data & AP & AP$_r$ & AP$_c$ & AP$_f$\\
    \hline
    O365 &  23.5 & 16.2 & 21.1 & 27.0\\
    O365,GQA & 31.9 & 22.5 & 29.9 & 35.4 \\
    O365,GoldG & 32.5 & 22.3 & 30.6 & \tb{36.0} \\
    O365,GoldG,CC3M$^\dagger$ & \tb{33.0} & \tb{23.6} & \tb{32.0}  & 35.5 \\
    \bottomrule
    \end{tabular}
    \caption{\textbf{Ablations on Pre-training Data.} We evaluate the zero-shot performance on LVIS of pre-training \name{} with different amounts of data.}
    \vspace{-10pt}
    \label{tab:ablation_pretrain_data}
\end{table}

\paragraph{Ablations on RepVL-PAN.} 
Tab.~\ref{tab:ablation_vl_pan} demonstrates the effectiveness of the proposed RepVL-PAN of \name{}, including Text-guided CSPLayers and Image Pooling Attention, for the zero-shot LVIS detection.
Specifically, we adopt two settings, \ie, (1) pre-training on O365 and (2) pre-training on O365 \& GQA.
Compared to O365 which only contains category annotations, GQA  includes rich texts, particularly in the form of noun phrases.
As shown in Tab.~\ref{tab:ablation_vl_pan}, the proposed RepVL-PAN improves the baseline (YOLOv8-PAN~\cite{yolov8_ultralytics}) by 1.1 AP on LVIS, and the improvements are remarkable in terms of the rare categories (AP$_r$) of LVIS, which are hard to detect and recognize.
In addition, the improvements become more significant when \name{} is pre-trained with the GQA dataset and experiments indicate that the proposed RepVL-PAN works better with rich textual information.

\begin{table}[]
    \centering
    \renewcommand\arraystretch{1.2}
    \begin{tabular}{ccc cccc}
    \toprule
    GQA & T$\rightarrow$I & I$\rightarrow$T & AP & AP$_r$ & AP$_c$ & AP$_f$\\
    \hline
    \xmark & \xmark & \xmark & 22.4 & 14.5 & 20.1 & 26.0 \\
    \xmark & \checkmark & \xmark  & 23.2 & 15.2 & 20.6 & 27.0 \\
    \xmark & \checkmark & \checkmark & 23.5 & 16.2 & 21.1 & 27.0 \\
    \hline
    \checkmark & \xmark & \xmark & 29.7 & 21.0 & 27.1 & 33.6 \\
    \checkmark & \checkmark & \checkmark & \tb{31.9} & \tb{22.5} & \tb{29.9} & \tb{35.4} \\
    \bottomrule
    \end{tabular}
    \caption{\textbf{Ablations on Re-parameterizable Vision-Language Path Aggregation Network.} We evaluate the zero-shot performance on LVIS of the proposed Vision-Language Path Aggregation Network. T$\rightarrow$I and I$\rightarrow$T denote the Text-guided CSPLayers and Image-Pooling Attention, respectively.}
    \label{tab:ablation_vl_pan}
\end{table}

\paragraph{Text Encoders.}
In Tab.~\ref{tab:ablation_text_encoder}, we compare the performance of using different text encoders, \ie, BERT-base~\cite{BERT} and CLIP-base (ViT-base)~\cite{CLIP}.
We exploit two settings during pre-training, \ie, frozen and fine-tuned, and the learning rate for fine-tuning text encoders is a $0.01\times$ factor of the basic learning rate.
As Tab.~\ref{tab:ablation_text_encoder} shows, the CLIP text encoder obtains superior results than BERT (+10.1 AP for rare categories in LVIS), which is pre-trained with image-text pairs and has better capability for vision-centric embeddings.
Fine-tuning BERT during pre-training brings significant improvements (+3.7 AP) while fine-tuning CLIP leads to a severe performance drop.
We attribute the drop to that fine-tuning on O365 may degrade the generalization ability of the pre-trained CLIP, which contains only 365 categories and lacks abundant textual information.

\begin{table}[]
    \centering
    \renewcommand\arraystretch{1.2}
    \begin{tabular}{ll cccc}
    \toprule
    Text Encoder & Frozen? & AP & AP$_r$ & AP$_c$ & AP$_f$\\
    \hline
    BERT-base & Frozen & 14.6 & 3.4 & 10.7 & 20.0 \\
    BERT-base & Fine-tune & 18.3 & 6.6 & 14.6 & 23.6 \\
    \hline
    CLIP-base & Frozen & \tb{22.4} & \tb{14.5} & \tb{20.1} & \tb{26.0} \\
    CLIP-base & Fine-tune & 19.3 & 8.6 & 15.7 & 24.8 \\
    \bottomrule
    \end{tabular}
    \caption{\textbf{Text Encoder in \name{}.} We ablate different text encoders in \name{} through the zero-shot LVIS evaluation.}
    \label{tab:ablation_text_encoder}
\end{table}

\subsection{Fine-tuning \name{}}
In this section, we further fine-tune \name{} for close-set object detection on the COCO dataset and LVIS dataset to demonstrate the effectiveness of the pre-training.

\paragraph{Experimental Setup.}
We use the pre-trained weights to initialize \name{} for fine-tuning.
All models are fine-tuned for 80 epochs with the AdamW optimizer and the initial learning rate is set to 0.0002.
In addition, we fine-tune the CLIP text encoder with a learning factor of 0.01.
For the LVIS dataset, we follow previous works~\cite{ViLD,DetPro,Detic} and fine-tune \name{} on the LVIS-base (common \& frequent) and evaluate it on the LVIS-novel (rare).
\paragraph{COCO Object Detection.}
We compare the pre-trained \name{} with previous YOLO detectors~\cite{YOLOv6,YOLOv7,yolov8_ultralytics} in Tab.~\ref{tab:finetune_coco_yolo}.
For fine-tuning \name{} on the COCO dataset, we remove the proposed RepVL-PAN for further acceleration considering that the vocabulary size of the COCO dataset is small.
In Tab.~\ref{tab:finetune_coco_yolo}, it's evident that our approach can achieve decent zero-shot performance on the COCO dataset, which indicates that \name{} has strong generalization ability.
Moreover, \name{} after fine-tuning on the COCO \texttt{train2017} demonstrates higher performance compared to previous methods trained from scratch.

\begin{table}[]
    \centering
    \renewcommand\arraystretch{1.2}
    \setlength{\tabcolsep}{5pt}
    \scalebox{0.95}{\begin{tabular}{lc ccc c}
    \toprule
    Method & Pre-train & AP & AP$_{50}$ & AP$_{75}$ & FPS \\
    \hline
    \multicolumn{6}{l}{\textit{Training from scratch.}} \\
    YOLOv6-S~\cite{YOLOv6} & \xmark & 43.7 & 60.8 & 47.0 & 442 \\
    YOLOv6-M~\cite{YOLOv6} & \xmark & 48.4 & 65.7 & 52.7 & 277 \\
    YOLOv6-L~\cite{YOLOv6} & \xmark & 50.7 & 68.1 & 54.8 & 166 \\
    YOLOv7-T~\cite{YOLOv7} & \xmark & 37.5 & 55.8 & 40.2 & 404 \\
    YOLOv7-L~\cite{YOLOv7} & \xmark & 50.9 & 69.3 & 55.3 & 182 \\
    YOLOv7-X~\cite{YOLOv7} & \xmark & 52.6 & 70.6 & 57.3 & 131 \\
    YOLOv8-S~\cite{yolov8_ultralytics} & \xmark & 44.4 & 61.2 & 48.1 & 386 \\
    YOLOv8-M~\cite{yolov8_ultralytics} & \xmark & 50.5 & 67.3 & 55.0 & 238 \\
    YOLOv8-L~\cite{yolov8_ultralytics} & \xmark & 52.9 & 69.9 & 57.7 & 159 \\
    \hline
    \multicolumn{6}{l}{\textit{Zero-shot transfer.}} \\
    \name{}-S & O+G & 37.6 & 52.3 & 40.7 & - \\
    \name{}-M & O+G & 42.8 & 58.3 & 46.4 & - \\
    \name{}-L & O+G & 44.4 & 59.8 & 48.3 & - \\
    \name{}-L & O+G+C & 45.1 & 60.7 & 48.9 & - \\
    \hline
    \multicolumn{6}{l}{\textit{Fine-tuned w/ RepVL-PAN.}} \\
    \name{}-S & O+G & 45.9 & 62.3 & 50.1 & - \\
    \name{}-M & O+G & 51.2 & 68.1 & 55.9 & - \\
    \name{}-L & O+G+C & 53.3 & 70.1 & 58.2 & - \\
    \hline
    \multicolumn{6}{l}{\textit{Fine-tuned w/o RepVL-PAN.}} \\
    \name{}-S & O+G & 45.7 & 62.3 & 49.9 & 373 \\
    \name{}-M & O+G & 50.7 & 67.2 & 55.1 & 231 \\
    \name{}-L & O+G+C & 53.3 & 70.3 & 58.1 & 156 \\
    \bottomrule
    \end{tabular}}
    \caption{\textbf{Comparison with YOLOs on COCO Object Detection.} We fine-tune the \name{} on COCO \texttt{train2017} and evaluate on COCO \texttt{val2017}. The results of YOLOv7~\cite{YOLOv7} and YOLOv8~\cite{yolov8_ultralytics} are obtained from MMYOLO~\cite{mmyolo2022}. `O', `G', and `C' denote pertaining using Objects365, GoldG, and CC3M$^{\dagger}$, respectively. The FPS is measured on one NVIDIA V100 w/ TensorRT.}
    \label{tab:finetune_coco_yolo}
\end{table}

\paragraph{LVIS Object Detection.}
In Tab.~\ref{tab:finetune_lvis_ovd}, we evaluate the fine-tuning performance of \name{} on the standard LVIS dataset.
Firstly, compared to the oracle YOLOv8s~\cite{yolov8_ultralytics} trained on the full LVIS datasets, \name{} achieves significant improvements, especially for larger models, \eg, \name{}-L outperforms YOLOv8-L by 7.2 AP and 10.2 AP$_r$.
The improvements can demonstrate the effectiveness of the proposed pre-training strategy for large-vocabulary detection.
Moreover, \name{}, as an efficient one-stage detector, outperforms previous state-of-the-art two-stage methods~\cite{ViLD,Detic,FVLM,DetPro,BARON} on the overall performance without extra designs, \eg, learnable prompts~\cite{DetPro} or region-based alginments~\cite{ViLD}.

\begin{table}[]
    \centering
    \renewcommand\arraystretch{1.2}
    \setlength{\tabcolsep}{7pt}
    \scalebox{0.95}{\begin{tabular}{ll cccc}
    \toprule
    Method & AP & AP$_r$ & AP$_c$ & AP$_f$ \\
    \hline
    ViLD~\cite{ViLD} & 27.8 & 16.7 & 26.5 & 34.2 \\
    RegionCLIP~\cite{RegionCLIP} & 28.2 & 17.1 & - & -\\
    Detic~\cite{Detic} & 26.8 & 17.8 & - & - \\
    FVLM~\cite{FVLM} & 24.2 & 18.6 & - & - \\
    DetPro~\cite{DetPro} & 28.4 & 20.8 & 27.8 & 32.4 \\
    BARON~\cite{BARON} & 29.5 & 23.2 & 29.3 & 32.5 \\
    \hline
    \textcolor{gray}{YOLOv8-S} & \textcolor{gray}{19.4} & \textcolor{gray}{7.4} & \textcolor{gray}{17.4} & \textcolor{gray}{27.0}\\
    \textcolor{gray}{YOLOv8-M} & \textcolor{gray}{23.1} & \textcolor{gray}{8.4} & \textcolor{gray}{21.3} & \textcolor{gray}{31.5} \\
    \textcolor{gray}{YOLOv8-L} & \textcolor{gray}{26.9} & \textcolor{gray}{10.2} & \textcolor{gray}{25.4} & \textcolor{gray}{35.8} \\
    \hline
    \name{}-S & 23.9 & 12.8 & 20.4 & 32.7 \\
    \name{}-M & 28.8 & 15.9 & 24.6 & 39.0\\
    \name{}-L & 34.1 & 20.4 & 31.1 & 43.5 \\
    \bottomrule
    \end{tabular}}
    \caption{\textbf{Comparison with Open-Vocabulary Detectors on LVIS.} We train \name{} on the LVIS-base (including common and frequent) report the \textit{bbox AP}. The \textcolor{gray}{YOLO-v8} are trained on the full LVIS datasets (including base and novel) along with the class balanced sampling.}
    \label{tab:finetune_lvis_ovd}
\end{table}

\subsection{Open-Vocabulary Instance Segmentation}
\label{ovis}
In this section, we further fine-tune \name{} for segmenting objects under the open-vocabulary setting, which can be termed open-vocabulary instance segmentation (OVIS).
Previous methods~\cite{OVIS} have explored OVIS with pseudo-labelling on novel objects.
Differently, considering that \name{} has strong transfer and generalization capabilities, we directly fine-tune \name{} on a subset of data with mask annotations and evaluate the segmentation performance under large-vocabulary settings.
Specifically, we benchmark open-vocabulary instance segmentation under two settings:
\begin{itemize}
    \item (1) \textit{COCO to LVIS} setting, we fine-tune \name{} on the COCO dataset (including 80 categories) with mask annotations, under which the models need to transfer from 80 categories to 1203 categories ($80\rightarrow{}1203$);
    \item (2) \textit{LVIS-base to LVIS} setting, we fine-tune \name{} on the LVIS-base (including 866 categories, common \& frequent) with mask annotations, under which the models need to transfer from 866 categories to 1203 categories ($866\rightarrow{}1203$).
\end{itemize}
We evaluate the fine-tuned models on the standard LVIS \texttt{val2017} with 1203 categories, in which 337 rare categories are unseen and can be used to measure the open-vocabulary performance.

\paragraph{Results.}
Tab.~\ref{tab:ovis_lvis} shows the experimental results of extending \name{} for open-vocabulary instance segmentation.
Specifically, we adopt two fine-tuning strategies: (1) only fine-tuning the segmentation head and (2) fine-tuning all modules.
Under strategy (1), the fine-tuned \name{} still retains the zero-shot capabilities acquired from the pre-training stage, allowing it to generalize to unseen categories without additional fine-tuning. 
Strategy (2) enables \name{} fit the LVIS dataset better, but it may result in the degradation of the zero-shot capabilities.

Tab.~\ref{tab:ovis_lvis} shows the comparisons of fine-tuning \name{} with different settings (COCO or LVIS-base) and different strategies (fine-tuning seg. head or fine-tuning all).
Firstly, fine-tuning on LVIS-base obtains better performance compared to that based on COCO.
However, the ratios between AP and AP$_r$ (AP$_r$/AP) are nearly unchanged, \eg, the ratios of \name{} on COCO and LVIS-base are 76.5\% and 74.3\%, respectively.
Considering that the detector is frozen, we attribute the performance gap to the fact that the LVIS dataset provides more detailed and denser segmentation annotations, which are beneficial for learning the segmentation head.
When fine-tuning all modules, \name{} obtains remarkable improvements on LVIS, \eg, \name{}-L achieves 9.6 AP gain.
However, the fine-tuning might degrade the open-vocabulary performance and lead to a 0.6 box AP$_r$ drop for \name{}-L.

\begin{table*}[h]
    \centering
    \renewcommand\arraystretch{1.2}
    {\begin{tabular}{llc cccc cc}
    \toprule
    Model & Fine-tune Data & Fine-tune Modules & AP & AP$_r$ & AP$_c$ & AP$_f$ & AP$^b$& AP$^b_r$ \\
    \hline
    \name{}-M & COCO & \textit{Seg Head} & 12.3 & 9.1 & 10.9 & 14.6 & 22.3 & 16.2\\
    \name{}-L & COCO & \textit{Seg Head} & 16.2 & 12.4 & 15.0 & 19.2 & 25.3 & \tb{18.0} \\
    \hline
    \name{}-M & LVIS-base & \textit{Seg Head} & 16.7 & 12.6 & 14.6 & 20.8 & 22.3 & 16.2\\
    \name{}-L & LVIS-base & \textit{Seg Head} & 19.1 & 14.2 & 17.2 & 23.5 & 25.3 & \tb{18.0}\\
    \hline
    \name{}-M & LVIS-base & \textit{All} & 25.9 & 13.4 & 24.9 & 32.6 & 32.6 & 15.8 \\
    \name{}-L & LVIS-base & \textit{All} & \tb{28.7} & \tb{15.0} & \tb{28.3} & \tb{35.2} & \tb{36.2} & 17.4 \\
    \bottomrule
    \end{tabular}}
    \caption{\textbf{Open-Vocabulary Instance Segmentation.} We evaluate \name{} for open-vocabulary instance segmentation under the two settings. We fine-tune the segmentation head or all modules of \name{} and report \textit{Mask AP} for comparison. AP$^b$ denotes the box AP.}
    \label{tab:ovis_lvis}
\end{table*}

\subsection{Visualizations}
\begin{figure*}[h]
    \centering
    \includegraphics[width=\linewidth]{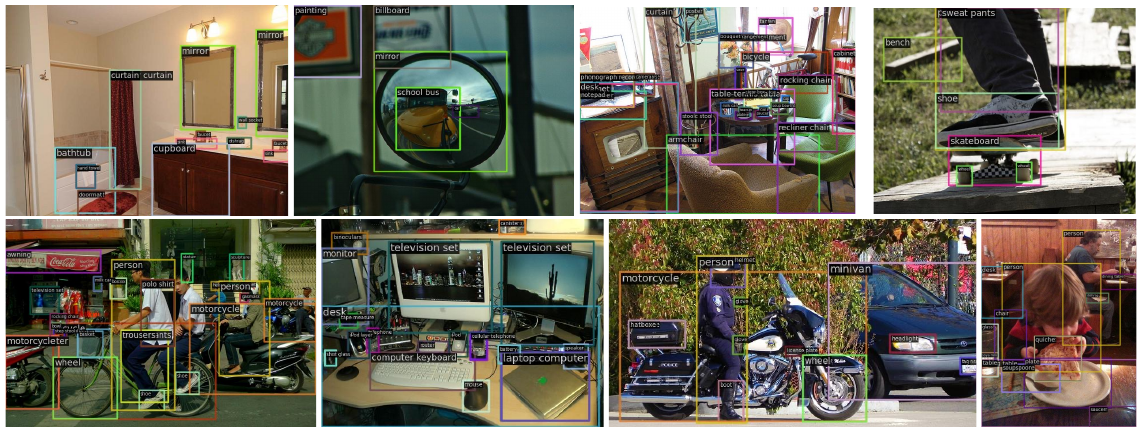}
    \caption{\textbf{Visualization Results on Zero-shot Inference on LVIS.} We adopt the pre-trained \name{}-L and infer with the LVIS vocabulary (containing 1203 categories) on the COCO \texttt{val2017}.}
    \label{fig:add_vis_lvis}
\end{figure*}

We provide the visualization results of pre-trained \name{}-L under three settings: (a) we perform zero-shot inference with LVIS categories; (b) we input the custom prompts with fine-grained categories with attributes; (c) referring detection.
The visualizations also demonstrate that \name{} has a strong generalization ability for open-vocabulary scenarios along with referring ability.
\paragraph{Zero-shot Inference on LVIS.}
Fig.~\ref{fig:add_vis_lvis} shows the visualization results based on the LVIS categories which are generated by the pre-trained \name{}-L in a zero-shot manner.
The pre-trained \name{} exhibits strong zero-shot transfer capabilities and is able to detect as many objects as possible within the image.

\begin{figure*}[h]
    \centering
    \includegraphics[width=\linewidth]{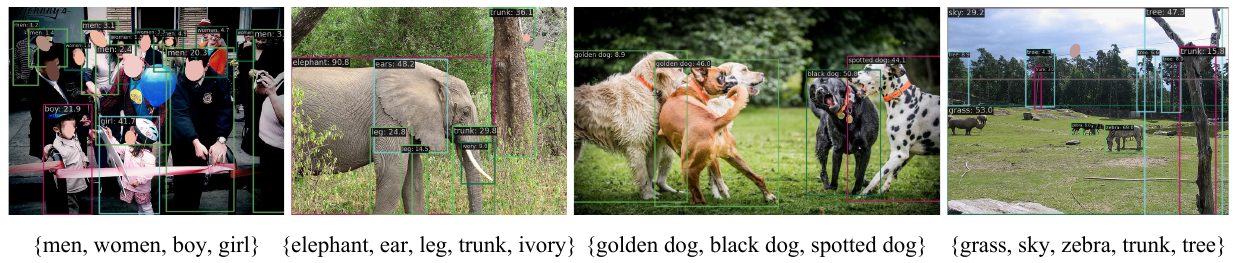}
    \caption{\textbf{Visualization Results on User's Vocabulary.} We define the custom vocabulary for each input image and \name{} can detect the accurate regions according to the vocabulary. Images are obtained from COCO \texttt{val2017}.}
    \label{fig:add_vis_user}
\end{figure*}

\paragraph{Inference with User's Vocabulary.}
In Fig.~\ref{fig:add_vis_user}, we explore the detection capabilities of \name{} with our defined categories.
The visualization results demonstrate that the pre-trained \name{}-L also exhibits the capability for (1) fine-grained detection (\ie, detect the parts of one object) and (2) fine-grained classification (\ie, distinguish different sub-categories of objects.).

\begin{figure*}[h]
    \centering
    \includegraphics[width=\linewidth]{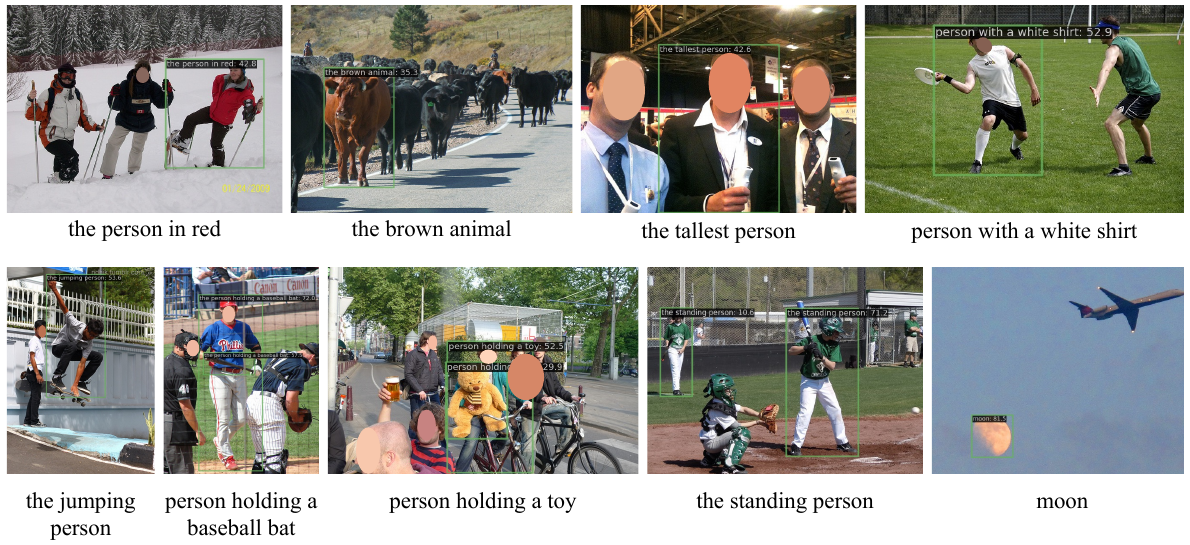}
    \caption{\textbf{Visualization Results on Referring Object Detection.} We explore the capability of the pre-trained \name{} to detect objects with descriptive noun phrases. Images are obtained from COCO \texttt{val2017}.}
    \label{fig:add_vis_refer}
\end{figure*}

\paragraph{Referring Object Detection.}
In Fig.~\ref{fig:add_vis_refer}, we leverage some descriptive (discriminative) noun phrases as input, \eg, the standing person, to explore whether the model can locate regions or objects in the image that match our given input.
The visualization results display the phrases and their corresponding bounding boxes, demonstrating that the pre-trained \name{} has the referring or grounding capability.
This ability can be attributed to the proposed pre-training strategy with large-scale training data.

\section{Conclusion}

We present \name{}, a cutting-edge real-time open-vocabulary detector aiming to improve efficiency and open-vocabulary capability in real-world applications.
In this paper, we have reshaped the prevalent YOLOs as a vision-language YOLO architecture for open-vocabulary pre-training and detection and proposed RepVL-PAN, which connects vision and language information with the network and can be re-parameterized for efficient deployment.
We further present the effective pre-training schemes with detection, grounding and image-text data to endow \name{} with a strong capability for open-vocabulary detection.
Experiments can demonstrate the superiority of \name{} in terms of speed and open-vocabulary performance and indicate the effectiveness of vision-language pre-training on small models, which is insightful for future research.
We hope \name{} can serve as a new benchmark for addressing real-world open-vocabulary detection.

{
    \small
    \bibliographystyle{ieeenat_fullname}
    \bibliography{main}
}

\appendix

\section{Additional Details}

\subsection{Re-parameterization for RepVL-PAN}
During inference on an offline vocabulary, we adopt re-parameterization for RepVL-PAN for faster inference speed and deployment.
Firstly, we pre-compute the text embeddings $W\in \mathbb{R}^{C\times D}$ through the text encoder.

\paragraph{Re-parameterize T-CSPLayer.}
For each T-CSPLayer in RepVL-PAN, we can re-parameterize and simplify the process of adding text guidance by reshaping the text embeddings $W\in \mathbb{R}^{C\times D\times1\times1}$ into the weights of a $1\times1$ convolution layer (or a linear layer), as follows:
\begin{equation}
    X' = X \odot \texttt{Sigmoid}(\texttt{max}(\texttt{Conv}(X, W), \texttt{dim=1})),
\end{equation}
where $X\times \in \mathbb{R}^{B\times D\times H\times W}$ and $X'\in \mathbb{R}^{B\times D\times H\times W}$ are the input and output image features. $\odot$ is the matrix multiplication with \texttt{reshape} or \texttt{transpose}.

\paragraph{Re-parameterize I-Pooling Attention.}
The I-Pooling Attention can be re-parameterize or simplified by:
\begin{equation}
    \tilde{X} = \texttt{cat}(\texttt{MP}(X_3, 3),\texttt{MP}(X_4, 3), \texttt{MP}(X_5, 3)),
\end{equation}
where \texttt{cat} is the concentration and \texttt{MP}($\cdot$, 3) denotes the max pooling for $3\times3$ output features. $\{X_3, X_4, X_5\}$ are the multi-scale features in RepVL-PAN.
$\tilde{X}$ is flattened and has the shape of $B\times D\times27$.
Then we can update the text embeddings by:
\begin{equation}
    W' = W + \texttt{Softmax}(W\odot\tilde{X}), \texttt{dim=-1}) \odot W,
\end{equation}

\subsection{Fine-tuning Details.}
We remove all T-CSPLayers and Image-Pooling Attention in RepVL-PAN when transferring \name{} to COCO~\cite{COCO} object detection, which only contains 80 categories and has a relatively low dependency on visual-language interaction.
During fine-tuning, we initialize \name{} using pre-trained weights.
The learning rate of fine-tuning is set to 0.0002 with the weight decay set to 0.05.
After fine-tuning, we pre-compute the class text embeddings with given COCO categories and store the embeddings into the weights of the classification layers.

\section{Automatic Labeling on Large-scale Image-Text Data}
\label{sec:autolabeling}
\begin{figure*}
    \centering
    \includegraphics[width=0.68\linewidth]{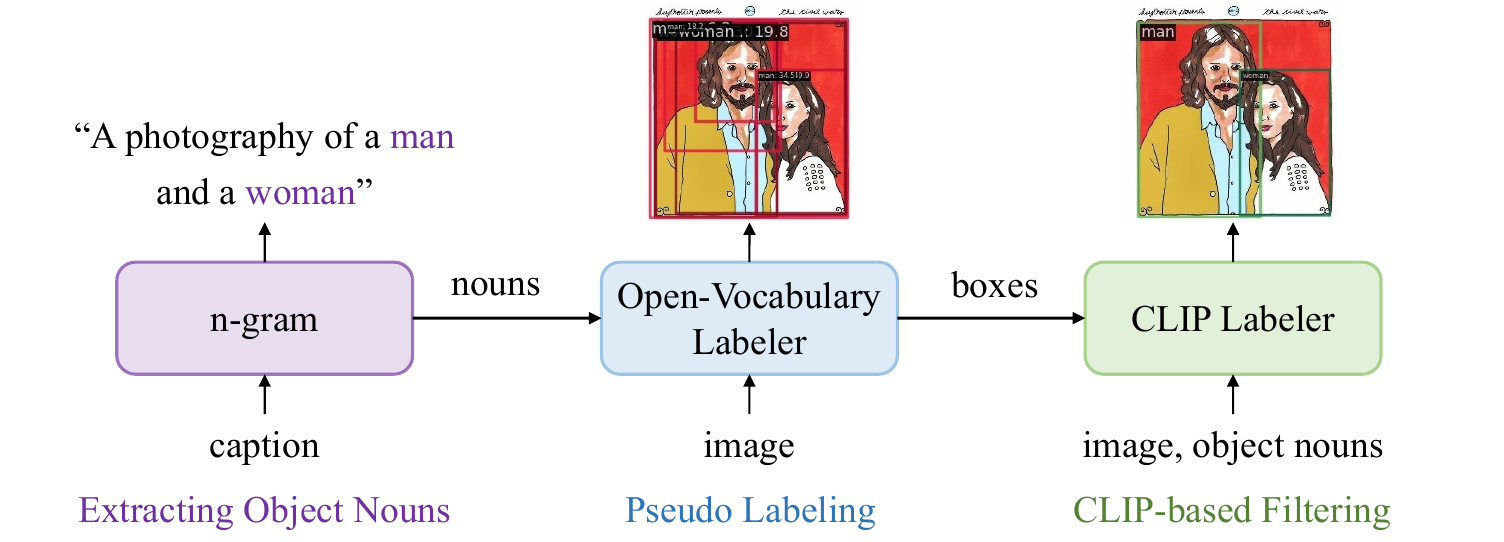}
    \caption{\textbf{Labeling Pipeline for Image-Text Data} We first leverage the simple n-gram to extract object nouns from the captions. We adopt a pre-trained open-vocabulary detector to generate pseudo boxes given the object nouns, which forms the coarse region-text proposals. Then we use a pre-trained CLIP to rescore or relabel the boxes along with filtering.}
    \label{fig:autolabel}
\end{figure*}

In this section, we add details procedures for labeling region-text pairs with large-scale image-text data, \eg, CC3M~\cite{CC3M}.
The overall labeling pipeline is illustrated in Fig.~\ref{fig:autolabel}, which mainly consists of three procedures, \ie, (1) extract object nouns, (2) pseudo labeling, and (3) filtering.
As discussed in \refmain{Sec.~3.4}, we adopt the simple n-gram algorithm to extract nouns from captions.

\paragraph{Region-Text Proposals.}
After obtaining the set of object nouns $T=\{t_k\}^K$ from the first step, we leverage a pre-trained open-vocabulary detector, \ie, GLIP-L~\cite{GLIP}, to generate pseudo boxes $\{B_i\}$ along with confidence scores $\{c_i\}$:
\begin{equation}
    \{B_i, t_i, c_i\}_{i=1}^N = \texttt{GLIP-Labeler}(I, T),
\end{equation}
where $\{B_i, t_i, c_i\}_{i=1}^N$ are the coarse region-text proposals.

\paragraph{CLIP-based Re-scoring \& Filtering.}
Considering the region-text proposals containing much noise, we present a restoring and filtering pipeline with the pre-trained CLIP~\cite{CLIP}.
Given the input image $I$, caption $T$, and the coarse region-text proposals $\{B_i, t_i, c_i\}_{i=1}^N$, the specific pipeline is listed as follows:
\begin{itemize}
    \item (1) Compute Image-Text Score: we forward the image $I$ with its caption $T$ into CLIP and obtain the image-text similarity score $s^{img}$. 
    \item (2) Compute Region-Text Score: we crop the region images from the input image according to the region boxes $\{B_i\}$. Then we forward the cropped images along with their texts $\{t_i\}$ into CLIP and obtain the region-text similarity $S^{r}=\{s^r_i\}_{i=1}^N$.
    \item (3) \texttt{[Optional]} Re-Labeling: we can forward each cropped image with all nouns and assign the noun with maximum similarity, which can help correct the texts wrongly labeled by GLIP.
    \item (4) Rescoring: we adopt the region-text similarity $S^r$ to rescore the confidence scores $\tilde{c_i} = \sqrt{c_i * s^r_i}$.
    \item (5) Region-level Filtering: we first divide the region-text proposals into different groups according to the texts and then perform non-maximum suppression (NMS) to filter the duplicate predictions (the NMS threshold is set to 0.5). Then we filter out the proposals with low confidence scores (the threshold is set to 0.3).
    \item (6) Image-level Filtering: we compute the image-level region-text scores $s^{region}$ by averaging the kept region-text scores. Then we obtain the image-level confidence score by $s=\sqrt{s^{img}*s^{region}}$ and we keep the images with scores larger than 0.3.
\end{itemize}
The thresholds mentioned above are empirically set according to the part of labeled results and the whole pipeline is automatic without human verification. Finally, the labeled samples are used for pre-training \name{}. We will provide the pseudo annotations of CC3M for further research.

\section{Pre-training \name{} at Scale}
When pre-training small models, \eg, \name{}-S, a natural question we have is: how much capacity does a small model have, and how much training data or what kind of data does a small model need?
To answer this question, we leverage different amounts of pseudo-labeled region-text pairs to pre-train \name{}.
As shown in Tab.~\ref{tab:zero_shot_yolos}, adding more image-text samples can increase the zero-shot performance of \name{}-S.
Tab.~\ref{tab:zero_shot_yolos} indicates: (1) adding image-text data can improve the overall zero-shot performance of \name{}-S; (2) using an excessive amount of pseudo-labeled data may have some negative effects for small models (\name{}-S), though it can improve the on rare categories (AP$_r$).
However, using fine-grained annotations (GoldG) for small models can provide significant improvements, which indicates that large-scale high-quality annotated data can significantly enhance the capabilities of small models.
And Tab.~\ref{tab:ablation_pretrain_data} in the main text has shown that pre-training with the combination of fine-annotated data and pseudo-annotated data can perform better. We will explore more about the data for pre-training small models or YOLO detectors in future work.

\begin{table*}[h]
    \centering
    \renewcommand\arraystretch{1.2}
    {\begin{tabular}{llc cccc}
    \toprule
    Method & Pre-trained Data & Samples & AP & AP$_r$ & AP$_c$ & AP$_f$\\
    \hline
    \name{}-S & O365 & 0.61M & 16.3 & 9.2 & 14.1 & 20.1 \\
    \name{}-S & O365+GoldG & 1.38M & 24.2 & 16.4 & 21.7 & 27.8 \\
    \hline
    \name{}-S & O365+CC3M-245k & 0.85M & 16.5 & 10.8 & 14.8 & 19.1 \\
    \name{}-S & O365+CC3M-520k & 1.13M & 19.2 & 10.7 & 17.4 & 22.4 \\
    \name{}-S & O365+CC3M-750k & 1.36M & 18.2 & 11.2 & 16.0 & 21.1 \\
    \bottomrule
    \end{tabular}}
    \vspace{-4pt}
    \caption{\textbf{Zero-shot Evaluation on LVIS.} We evaluate the performance of pre-training \name{}-S with different amounts of data, the image-text data.}
    \label{tab:zero_shot_yolos}
\end{table*}


\end{document}